\documentclass{ifacconf}

\usepackage{graphicx}      %
\usepackage{natbib}        %

\usepackage{amsfonts}
\usepackage{amsmath}
\usepackage{bbm} %

\usepackage{xcolor}

\newcommand{\R}{\mathbb{R}}
\newcommand{\E}{\mathbb{E}}
\newcommand{\V}{Var}
\DeclareMathOperator*{\argmin}{arg\,min}
\newcommand\norm[1]{\left\lVert#1\right\rVert}
\newcommand\newtag[2]{#1\def\@currentlabel{#1}\label{#2}}

\begin{document}
\begin{frontmatter}

\title{A novel Deep Neural Network architecture for non-linear system identification\thanksref{footnoteinfo}} 

 \thanks[footnoteinfo]{This project was partially supported by the Italian Ministry of University and Research under the PRIN'17 project "Data-driven learning of constrained control systems" , contract no. 2017J89ARP and  by NVIDIA Corporation trough the  GPU Grant Program.}
 
\author[First]{Luca Zancato} 
\author[Second]{Alessandro Chiuso} 

\address[First]{Department of Information Engineering, University of Padova, Padova 35131, Italy (e-mail: luca.zancato@phd.unipd.it).}
\address[Second]{Department of Information Engineering, University of Padova, Padova 35131, Italy (e-mail: chiuso@dei.unipd.it)}

\begin{abstract}                %
We present a novel Deep Neural Network (DNN) architecture for non-linear system identification. 
We foster generalization by constraining DNN representational power. To do so, inspired by fading memory systems, we introduce  inductive bias (on the architecture) and regularization (on the loss function). This architecture allows for automatic complexity selection based solely on available data, in this way the number of hyper-parameters that must be chosen by the user is reduced.
Exploiting the highly parallelizable DNN framework (based on Stochastic optimization methods) we successfully apply our method to large scale datasets.
\end{abstract}


\end{frontmatter}

\section{Introduction}

The main goal of system identification is to build a dynamical model from observed data,  which is of course expected to  generalize well on unseen data. In the context of non-linear systems, both parametric (see \cite{ljung1995nonlinear, juditskyBiasVariance, DeepAutoencoder}) and non-parametric models (see \cite{giapi}) are viable alternatives  used in practice. 
Recently many efforts have been devoted to extend classical results for linear systems to non-linear ones. Instances of parametric and non-parametric model classes are respectively NARX/NARMAX and Kernel based methods (e.g. see \cite{giapi}). 
Typically  the identification problem can be divided  in two steps: first find the best model class given the available data and then find the best  model within that particular model class. None of these two problems can be easily solved in general and often model optimization is a non-convex problem. %
Finding the proper model complexity (structure) requires a complexity criterion. 
Beyond classical complexity criteria such as AIC and BIC (see \cite{ljung1995nonlinear}) many other automatic model complexity criteria have been introduced, both in the parametric (see \cite{ANOVA}) and non-parametric frameworks (see \cite{giapi}).

The aim of this paper is to extend ideas proposed in \cite{giapi} to the parametric framework. More precisely we shall build a parametric estimator for non-linear system identification using  Neural Networks (NN) as building blocks.
Despite NNs' universal approximation property (see \cite{cybenko1989approximation}),  learnability is still by and large an open problem. Due to their high capacity, NNs are prone to overfitting unless constrained by  regularization or inductive bias (see \cite{rethinking}). As such, our 
 architecture and  optimization loss are specifically designed to exploit domain knowledge (fading memory systems) and to automatically detect and choose the best model complexity from training data. 
The inductive bias relies on the assumption that the system to be identified belongs to the class of fading memory systems  (\cite{fading_systems}).

Previously proposed non-parametric methods such as in \cite{giapi} might not scale well with the number of data; on the contrary, our architecture can scale to hundred of thousand datapoints (as is typically the case for NNs based models, see \cite{ljung1995nonlinear, DeepAutoencoder}). 
Furthermore, the parametric model and the loss function are designed so  that standard Deep Learning regularization techniques (\cite{orthonormalization, dropout, BN}) and Stochastic Optimization methods (\cite{adam, SGLD}) can be applied.

\subsection*{Notation}
In this paper capital letters $A$ will denote matrices, lowercase letters $a$ column vectors. The transpose of the matrix $A$ will be denoted with $A^\top$. Given a  time series
${y_t}$, $t\in \mathbb{Z}$, we shall denote with 
$y_t^- := [y_{t-1}, y_{t-2}, ...]^\top$ the infinite past, while $\varphi_{t,T}(y)$ will denote the finite past of length $T$, i.e. $\varphi_{t,T}(y) := [y_{t-1}, y_{t-2}, ..., y_{t - T}]^\top $.
The  Frobenius norm of a matrix $A$ will be denoted with $\norm{A}_F^2:= Tr(A^\top A)$, while for the weighted 2-norm of a vector $a$ we shall use the notation $\norm{a}^2_\Sigma := a^\top \Sigma a$.

\section{Problem Statement}
Let $\{u_t\}$ and $\{y_t\}$, $t \in \mathbb{Z}$  be respectively the input and output of a discrete time, time invariant,  nonlinear state-space stochastic system
\begin{equation}\label{eqn:nonlineramodel}
    \begin{array}{rcl}
    x_{t+1} & = & f(x_t,u_t,w_t) \\
    y_t & = & h(x_t) + v_t
    \end{array}
\end{equation}
where $\{w_t\}$ and $\{v_t\}$ are respectively process and measurement noises. A rather standard assumption is that both $\{w_t\}$ and $\{v_t\}$ are strictly white and independent. For ease of exposition we shall assume that both $y$ and $u$ are scalar, but extension to the vector case is straightforward. We will denote with $z_t:=(y_t,u_t)^\top$ the joint input-output process.

Defining the one-step-ahead predictor:
\begin{equation}\label{eqn:one-step-predictor}
    \hat{y}_{t|t-1} = F_0(z_t^-) := \E[y_t | z_t^-] 
\end{equation}  the input-output behaviour of the state space model \eqref{eqn:nonlineramodel} can be written in \emph{innovation} form as
\begin{equation}\label{eqn:innotation_form}
y_t = F_0(z_t^-) + e_t
\end{equation}
where $e_t$ is, by definition, the one step ahead prediction error (or  \emph{innovation sequence}) of $y_t$ given the joint past
$\{z_s, s< t\}$. The innovation  $e_t$ is a martingale difference sequence w.r.t. the sigma algebra generated by  past data 
${\cal P}_t: = \sigma\{z_s, s< t\} = \sigma\{y_t^-, u_t^-\}$ and, thanks to the time-invariance assumption on  \eqref{eqn:nonlineramodel},  it has constant conditional variance
\begin{equation}\label{eqn:innovation_moments}
    \V[e_t] = \V[y_t|z_t^-] = \eta^2, \quad t=1,...,N
\end{equation}
We shall also assume that $e_t$ is strictly white.

Our main goal is to find an estimate  $\hat F$ of the predictor map $F_0$ in \eqref{eqn:one-step-predictor}.  This problem can be framed in the classical regularized Prediction Error Method (PEM) framework, i.e. defining\footnote{W.l.o.g we use the square loss.}
$$
\hat F = {\rm arg\;\; min}_{F \in {\cal F}} \frac{1}{N} \sum_{t=1}^N (y_t - F(z_t^-))^2 + \lambda P(F)
$$
where ${\cal F}$ is the  model class and $P(F)$ is a penalty function. 

This framework includes both classical parametric approaches (i.e. where ${\cal F}$ is a parametric model class ${\cal F}:={F_W, W \in \R^k}$ and the penalty $P(F)$ is expressed as a function of the parameters $W$) as well as non-parametric ones where $F$ lives in an infinite dimensional space such as a Reproducing Kernel Hilbert space (RKHS) and $P(F)$ is the norm in the space. It is well known that under mild assumptitons solving a Tikhonov regularization problem in RKHS under the square loss is equivalent to MAP estimation in the frameworks of Gaussian Processes (GP) \cite{Rasmussen}. Thus we shall interchangeably refer to RKHS and GPs.

In particular, we shall compare state-of-the art nonparametric methods introduced in \cite{giapi} that use RKHS/GPs with a class of Neural Networks models that will be introduced in the next Section.

\section{Non-linear Model structures}

In this work, similarly to \cite{giapi},  we shall consider a class of non-linear systems also known as fading memory systems (see e.g. \cite{fading_systems} and references therein), a property that can be informally described by saying that the effect of past inputs $u_s$, $s\leq t$ on the output $y_t$ becomes negligible (tends to zero asymptotically) as $t-s$ goes to infinity. This property guarantees that the system behaviour can be uniformly approximated on compact sets. 

Therefore the universal approximation properties of Neural Networks (NN) (see e.g. \cite{cybenko1989approximation}) suggests that NNs can be seen as natural candidates to tackle the identification problem. 
Yet,  NNs are known to suffer from severe overfitting. To cure this limitation, we introduce: 
\begin{itemize}
    \item structure in the NN architecture (inductive bias)
    \item a suitable regularization on the Networks coefficients 
\end{itemize}
Both inductive bias and regularization are designed to encode the fading memory property described above. 

Under the fading memory assumption we shall assume that the predictor model 
$F(y_t^-,u_t^-)$ in \eqref{eqn:one-step-predictor} depends only upon a finite, yet arbitrarily long window of past data, i.e.
\begin{equation}\label{finite-memory}
F(y_t^-,u_t^-) = F(y_{t-1},u_{t-1},...,y_{t-T},u_{t-T}) = F(\varphi_{t,T}(z)) 
\end{equation}
The past horizon $T$ is finite but \emph{arbitrarily long} so that no significant bias is introduced.  Provided a suitable regularization is used, $T$ can be taken to be arbitrarily large and need not  perform a bias-variance tradeoff.

\subsection{Single Hidden Layer Feedforward Neural Networks}
The simplest possible structure is provided by the so-called
one-hidden-layer Feedforward Neural Network:
\begin{equation}\label{eqn:single_layer}
    f_{\mathbf{W}}(z)  = W_2 g(W_1\varphi_{t,T}(z) + b_1) + b_2
\end{equation}
where $W_1 \in \mathbb{R}^{n_1 \times n_0}$,  $b_1 \in \mathbb{R}^{n_1}$, $W_2 \in \mathbb{R}^{n_2 \times n_1}$, $b_2 \in \mathbb{R}^{n_2}$, in our case $n_0 = 2T$ and $n_2=1$ (since we are assuming scalar signals). The number of hidden units $n_1$  is a user choice and the activation function $g$ is typically a \emph{sigmoid}, a Rectified Linear Unit (ReLU)   or a smooth version of the latter known as Exponential Linear Unit (ELU).

\subsection{Multilayer Feedforward Neural Networks}
Multilayer Feedforward Neural Networks (or Deep Neural Networks DNNs) are a straightforward extension of the single layer network: they are achieved simply by stacking layers of non-linearities on top of the others: 
\begin{equation}\label{eqn:DNN}
     f_{\mathbf{W}}(z)  = W_L(W_{L-1}(...(W_1 \varphi_{t,T}(z) + b_1)...)+b_{L-1}) + b_{L}
\end{equation}
If we define $h_l$ the $l$-th hidden layer and $s_l$ the output of the $l$-th linear map we can write the following: 
\begin{equation}
    s_l = W_l h_{l-1} + b_l, \quad h_{l} = g(s_{l}) \quad l = 1,...,L-1
\end{equation}
and the output of the network is $s_L = W_L h_{L-1} + b_L$.
Note $h_0 = \varphi_{t,T}(z)$, $W_l \in \mathbb{R}^{n_l \times n_{l-1}}$, $n_0 = 2T$ and $n_L=1$.
Both for DNNs and single layer networks (L=2) we denote all their parameters as $\mathbf{W}=\{W_i, b_i\}$ for $i=1,...,L$, the total number of parameters is $\sum_{l = 1}^{L}(n_{l-1}+1)n_l$.

\section{Network Architecture}

Inspired by \cite{giapi}  we now introduce a block-structured architecture that can be used to encode the fading memory assumption. In the next section, using suitable design regularization schemes, we endow our model class with the ability to automatically trade-off model complexity with the available data by tuning a parameter that encodes how fast memory of the past fades away. 
In order to avoid degeneracy issues, we exploit a standard tool in Deep Neural Networks, namely batch normalization. 

We assume the predictor function $F_0$ can be written as a linear combination of (in principle) infinitely many elementary building blocks
$f_{\mathbf{W}_i}$, each of them described by a DNN. In particular we will assume that  each $f_{\mathbf{W}_i}$ is actually a function of only a small window of past data (of length $p$), namely:
\begin{align}\label{eqn:blocks_fading}
    f_{\mathbf{W}_i} :=& f_{\mathbf{W}_i}(y_{t-i-1},  u_{t-i-1}, ..., y_{t-i-p}, u_{t-i-p}) \\
                      =& f_{\mathbf{W}_i}(\varphi_{t-i,p} (z)) \in \R^c
\end{align}
where w.l.o.g. we are considering the same horizon $p$ both for the past of $y$ and $u$.

Note that each block $f_{\mathbf{W}_i}(\varphi_{t-i,p}(z)) $ outputs a feature vector of dimension $c$ that should be chosen when defining the block structure (e.g. one can choose $c=dim(y_t)$). 

The output predictor is then parametrized in the form 
$$F_{\theta, \mathbf{W}}=\sum_{i=0}^{\infty} \theta_i^\top f_{\mathbf{W}_i}$$

\begin{figure}
    \centering
        \includegraphics[width=8.5cm]{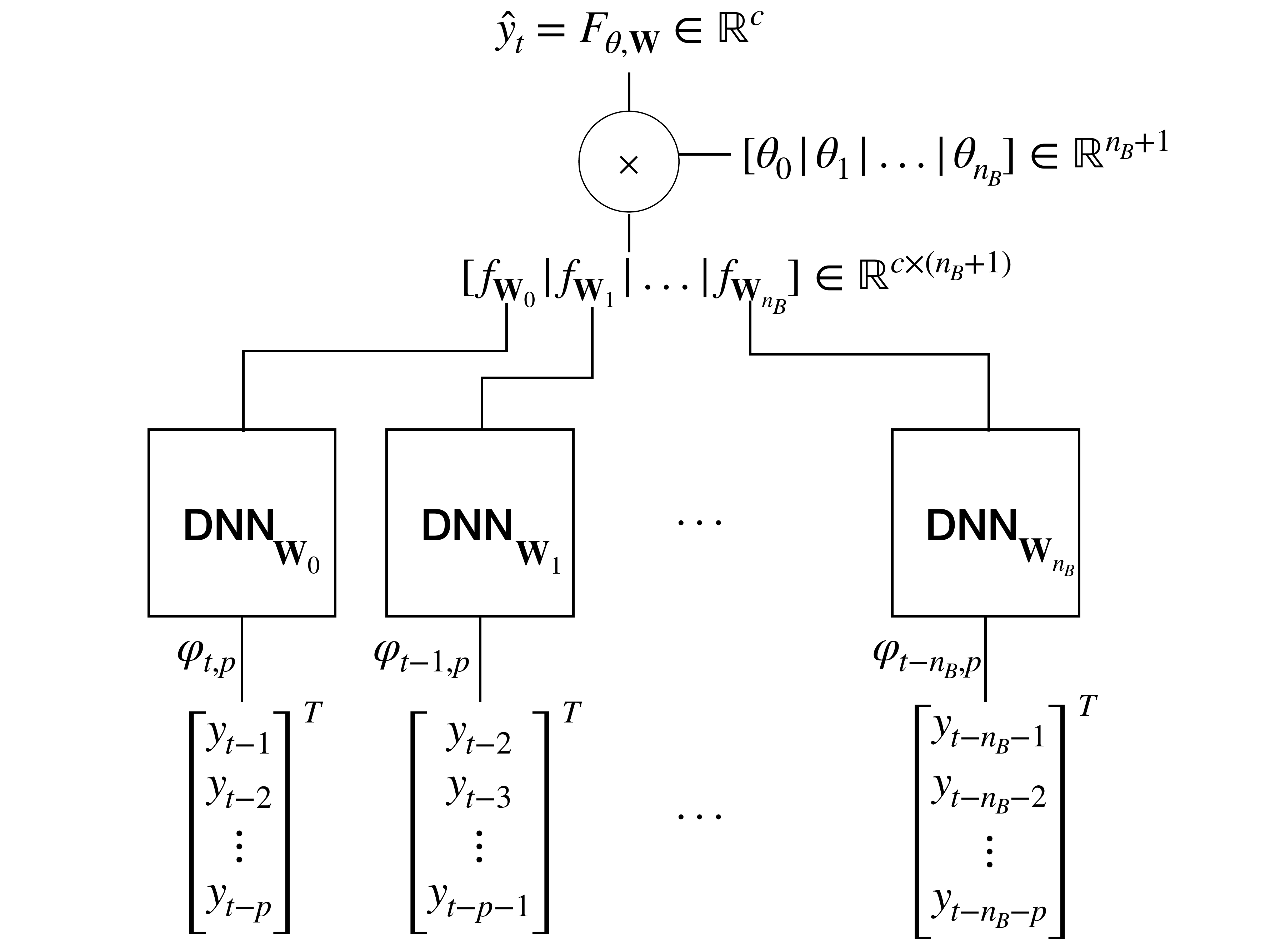}
        \caption{\textbf{Fading architecture.}}
        \label{fig:FadingArchitecture}
\end{figure}

For the sake of simplicity consider the case $c=1$, which means each block is processing a translated window of $p$ past measurements and outputs a single scalar. 
The fading memory assumption guarantees also that the contribution to output prediction of blocks $f_{\mathbf{W}_i}(\varphi_{t-i,p})$ should fade to zero 
with the index $i$. Thus, w.l.o.g., we shall consider a finite number of blocks $n_B +1$ and thus truncate the model $F_{\theta, \mathbf{W}}$ to the form
$$F_{\theta, \mathbf{W}}=\sum_{i=0}^{n_B} \theta_i^\top f_{\mathbf{W}_i}$$
as illustrated in Figure \ref{fig:FadingArchitecture}.

Ideally  $n_B$ should be large enough to capture the memory of the system, so that the network can approximate arbitrarily well  the ``true'' $F_0$ and should \emph{not} be chosen to face a bias-variance trade-off. Regularization shall be used to control the model complexity, by automatically assigning fading weights to the outputs of each block.

\begin{rem} The choice of $f_{\mathbf{W}_i}$ is completely arbitrary, e.g. it could be a single layer, multilayer or basis functions Neural Network; each  block has its own  set of parameters, so that it can potentially extract different features from different lagged past windows. 
\end{rem}

Overall the network is described by the parameters:
\begin{itemize}
\item $n_B$ (number of blocks)
\item p (size of ``elementary'' regressor $\varphi_{t-i,p}(z)$) 
\item the  block weights $\mathbf{W} = \{\mathbf{W}_0,\mathbf{W}_1,...,\mathbf{W}_{n_B}\}$ 
\item the recombination parameters $\theta_0$, $\theta_1$, ..., $\theta_{n_B}$.
\end{itemize}

\section{Fading Memory Regularization}\label{sec:fading_regularization}
In this section we introduce a regularized loss inspired by Bayesian arguments which allows us to use an architecture with a ``large enough''  number of blocks $n_{B}$ (i.e. larger than the actual system memory) and automatically select their weights to avoid overfitting.

We shall also assume that innovations \eqref{eqn:innotation_form}  are Gaussian, so that  $y_t|{\cal P}_t \sim \mathcal{N}(F(z_t^-), \eta^2)$. We   denote with $p(y_t|\theta, \mathbf{W}, {\cal P}_t )$ the conditional likelihood (Gaussian)  of $y_t$ given ${\cal P}_t $. 
In the following we shall denote with 
$Y:=[y_t, y_{t+1}, ..., y_{t+f-1}]^\top$ the 
set of outputs over which prediction is computed and with 
$\hat Y_{\theta,\mathbf{W}}$ the corresponding predictions parametrized by $\theta$ and ${\mathbf W}$ computed as in \eqref{eqn:one-step-predictor} with $F_0 = F_{\theta,\mathbf{W}}$.
The likelihood function takes the form
\begin{equation}\label{eqn:likelihood}
    p(Y|\theta, \mathbf{W}) = \prod_{k=0}^{f-1}p(y_{t+k}|\theta, \mathbf{W}, \varphi_{t+k, k}(z))   
\end{equation}
In a Bayesian framework the optimal parameter set can be found maximizing the posterior $p(\theta, \mathbf{W}| Y)$. Modeling $\theta$ and $\mathbf{W}$ as independent random variables we have: 
\begin{equation}\label{eqn:posterior}
    p(\theta, \mathbf{W}| Y) \propto p(Y|\theta, \mathbf{W}) p(\theta) p(\mathbf{W})
\end{equation}
where $p(\theta)$ is the prior associated to the fading coefficients and $p(\mathbf{W})$ is the prior on the parameters of the blocks.

In particular  $p(\theta)$ should reflect the fading memory assumption, e.g. assuming
$\theta_k$ have zero mean with exponentially decaying variances
$$
\E \theta_k^2  = \kappa \lambda^{k-1}.
$$
The maximum entropy prior $p_{\lambda, \kappa}(\theta)$ (see \cite{Cover}) under such constraints is 
\begin{equation}\label{eqn:fading_prior}
    \log(p_{\lambda, \kappa}(\theta)) \propto  -\norm{\theta}^2_{\Lambda^{-1}} - \log(|\Lambda|)
\end{equation}
where $\Lambda \in \mathbb{R}^{n_B+1}$ is a diagonal matrix with elements: $\Lambda_{i,i} = \kappa \lambda^{i-1}$ and $\kappa \in \R^+,  \lambda \in (0,1)$.

The parameter $\lambda$ represents how fast model output ``forgets" the past of $y$ and $u$. Therefore $\lambda$ regulates the complexity of $F_{\theta, \mathbf{W}}$: the smaller $\lambda$ the smaller the complexity. In practice we do not have access to this information and indeed we need to estimate $\lambda$ from data. 

One would be tempted to estimate jointly $\theta, \mathbf{W}, \lambda,\kappa$ (and possibly $\eta$) minimizing the negative log of  the joint posterior:
\begin{equation}\label{eqn:MAP}
    \argmin_{\theta, \mathbf{W},\lambda,\kappa} \frac{\norm{Y - \hat Y_{\theta, \mathbf{W}}}^2}{\eta^{2}} +  \log(\eta^2) - \log(p_{\lambda, \kappa}(\theta)) - \log(p(\mathbf{W}))
\end{equation}
Unfortunately this leads to a degeneracy, in that the joint negative log posterior goes to $-\infty$ when $\lambda \rightarrow 0$. 

Indeed typically the parameters describing the prior (such as $\lambda$) are estimated by maximizing the marginal likelihood, i.e. the likelihood of the data once the parameters ($\theta, \mathbf{W}$) have been integrated out. Unfortunately  the task of computing (or even approximating) the marginal likelihood in this setup is prohibitive and we should resort to Monte Carlo sampling techniques. While this is an avenue worth investigating, in this study we have preferred to adopt the following variational strategy inspired by the linear  setup. 

Indeed the model structure we consider is linear in $\theta$ and therefore we can write 
$$
\hat Y = F \theta$$ for a suitable defined 
$F$ built with the outputs of the blocks $f_{\mathbf{W}_i}$. Using this observation the following holds:
$$\begin{array}{rcl}
    \argmin_{\theta} \frac{1}{\eta^2}{\norm{Y - F \theta}^2}+  \theta^\top \Lambda^{-1} \theta &=& Y^\top \Sigma^{-1}Y 
\end{array}
$$
with $ \Sigma:=F\Lambda F^\top + \eta^2 I$. This guarantees that
$$
\frac{1}{\eta^2}{\norm{Y - F \theta}^2}+  \theta^\top \Lambda^{-1} \theta  + \log|{\Sigma}| \geq Y^\top \Sigma^{-1}Y  + \log|{\Sigma}|
$$ where the right hand side is (proportional to) the negative marginal likelihood with marginalization taken \emph{only} w.r.t. $\theta$.
Therefore 
$$
\frac{1}{\eta^2}{\norm{Y - \hat Y_{\theta,\mathbf{W}}}^2}+  \theta^\top \Lambda^{-1} \theta   +\log|{F\Lambda F^\top + \eta^2 I}|$$ is an upper bound of the marginal likelihood and does not suffer of the  degeneracy alluded before.

With this considerations in mind, and inserting back the optimization over ${\mathbf{W}}$, the overall optimization problem we solve is 
\begin{align}\label{eqn:MAP_rewritten}
    \argmin_{\theta, \mathbf{W}, \lambda \in (0,1), \kappa >0} &\frac{1}{\eta^2}\norm{Y_t - F_{\theta, \mathbf{W}}}^2 +  \log(p(\mathbf{W})) + \nonumber\\ 
    & + \norm{\theta}_{\Lambda^{-1}} + \log(|F\Lambda F^\top  + \eta^2 I|)
\end{align}

The last missing ingredient is the choice of the regularization on the block parametrization $\mathbf{W}$.
Two issues need to be accounted for: 

\begin{enumerate}
    \item  The output of each block should be properly  normalized to avoid degeneracy (non-identifiability) due to the multiplications $\theta_i f_{{\mathbf W}_i}$. To address this issue we resort to a standard tool used for Deep NN, namely batch normalization (see subsection \ref{block_norm}). 
    \item We should avoid that single blocks overfit and thus their complexity should be controlled (see subsection \ref{block_reg}). 

\end{enumerate}

\subsection{Normalization of the blocks}\label{block_norm}
As mentioned above, the blocks  $f_{\mathbf{W}_i}$ should be rich enough to model nonlinearities of the system, yet they should not undo the fading memory regularization we introduced. We can avoid degeneracy due to non-identifiability by properly normalizing the output of each block; we choose to apply a modern regularization method which is typically applied to regularize DNNs: Batch Normalization (see \cite{BN}). The main idea behind batch normalization is to maintain running statistics (means and standard deviations) of the outputs of the hidden nodes of a DNN model during training and apply a normalizing affine transformation to these outputs so that the inputs at each layer have zero mean and unit variance. In our case we do not want the output of each block to have zero mean and unit variance, rather we need comparable means and scales across each block output.
We therefore use batch normalization to normalize each block output and then we use an affine transformation (with parameters to be optimized) in order to jointly rescale all the output blocks together before the linear combination with $\theta$.

Denoting with $\bar{f}_{\mathbf{W}_i}$ the normalized $i$-th block output, the output of our regularized fading architecture is: $F_{\theta, \mathbf{W}} = \sum_{i=0}^{n_B} \theta_i \bar{f}_{\mathbf{W}_i}$. The normalization is performed according to:
\begin{equation}\label{eqn:batch_norm}
    \bar{f}_{\mathbf{W}_i} =  \frac{f_{\mathbf{W}_i} - \E[f_{\mathbf{W}_i}]}{\sqrt{\V[f_{\mathbf{W}_i}] + \epsilon_1}} \gamma + \beta \quad i=0,...,n_B
\end{equation}
where $\E[f_{\mathbf{W}_i}]$ and $\V[f_{\mathbf{W}_i}]$ are estimated using a running average along  the optimization iterations (as standard practice with batch normalization) and $\epsilon_1$ is a small number used to avoid numerical issues in case the estimated variance becomes too small. 

\begin{rem} $\gamma$ and $\beta$ are jointly optimized with other parameters and are shared among the outputs of the blocks such that the relative scale among them is preserved. 
\end{rem}

\subsection{Controlling block complexity}\label{block_reg}

Without regularization, each single block could overfit and therefore reduce generalization capabilities of our architecture.

As pointed out earlier batch normalization could be applied not only to the output layer of each block but it can also be applied layer-wise, it is well known that layer-wise batch norm improves trainability of DNNs, since it reduces the internal covariance shift (see \cite{BN}). Dropout is another commonly used method to reduce `neurons co-adaptation' and therefore improve generalization \cite{dropout}. In the following we shall mainly focus on another type of regularization which can directly be imposed following the Bayesian argument we used in \eqref{eqn:MAP_rewritten}: we shall impose a prior on $\mathbf{W}$ (for simplicity we consider each block $\mathbf{W}_i$, $i=0,...,n_B$ independently). 

Take now a single block $f_{\mathbf{W}_i}$, which is a DNN with $L$ layers, parametrized by $W_l$ and $b_l$ for $l=1,...,L$. 
Inspired by \cite{orthonormalization} we enforce that the Gram matrix of the weights matrix is close to the identity.
Therefore we consider the following per-layer regularization term: 
\begin{equation}\label{eqn:SOR}
    \log(W_l) = \norm{W_l^\top W_l - I_{n_l-1}}_F^2 \quad l = 1,...,L
\end{equation}
where $W_l \in \mathbb{R}^{n_l \times n_{l-1}}$. 

\begin{rem} We assume the priors are independent both across layers and across blocks (i.e. DNNs).
\end{rem}

Such a soft orthogonality regularization (SO) is known to foster network trainability by stabilizing the distribution of activations over layers \cite{orthonormalization}. 

\section{Optimization}
The optimization problem \eqref{eqn:MAP_rewritten} has been solved using off-the-shelf stochastic optimization tools such as Stochastic Gradient Descent (SGD) and Adam (see \cite{SGLD, adam}). Both these methods rely on gradients to find the best set of parameters, therefore we must require the fading architecture and its blocks to be differentiable w.r.t. their parameters (some extensions are applicable, e.g. with  ReLU activations functions).
Note the stochasticity introduced by the choice of the minibatches in SGD has been proven to be highly effective and provide properties which are not shared with Gradient Descent, such as the ability to avoid saddle points and spurious local minima of the loss function. 

\begin{rem} The stochasticity in the choice of the minibatches only affects the computation of the fit (minus log likelihood) term in \eqref{eqn:MAP_rewritten}  since  the regularization term does not need any datum to be computed.
\end{rem}

\section{Numerical Results}
\begin{table}
\begin{center}
\caption{Nonlinear systems (\cite{giapi}).} \label{tb:nlsys}
\begin{tabular}{ccll}
\hline
(\newtag{1}{eq:sys1}) &$y_t$ &=&    $e^{-0.1y_{t-1}^2}(2y_{t-1} - y_{t-2}) + e_t $\\
(\newtag{2}{eq:sys2}) &$y_t$ &=& $-2y_{t-1} \mathbf{1}(y_{t-1} <0) + 0.4y_{t-1} \mathbf{1}(y_{t-1} \geq 0) + e_t$ \\ 
(\newtag{3}{eq:sys3}) &$y_t$ &=& $0.5y_{t-1} -0.05y_{t-2}^2 + u^2_{t-1} + 0.8 u_{t-2} + 0.22e_t$ \\ 
(\newtag{4}{eq:sys4}) &$y_t$ &=&  $0.8 y_{t-1}+u_{t-1} - 0.3 u_{t-1}^3 + 0.25 u_{t-1}u_{t-2}$ \\
                &&&  $-0.3u_{t-2} +0.25u_{t-2}^3 -0.2u_{t-2}u_{t-3} -0.4u_{t-3}$ \\
                &&&  $+ 0.14e_t$ \\
\hline
\end{tabular}
\end{center}
\end{table}

Similarly to \cite{giapi} we tested our architecture using Monte Carlo studies on 4 nonlinear systems of increasing complexities, as  listed in Table \ref{tb:nlsys}.
For each nonlinear system we have generated random trajectories of length $N$ starting from the system initially at rest, we take $u_t \sim \mathcal{N}(0, 1)$ (whenever possible) and $e_t \sim \mathcal{N}(0, 1)$. 
We test generalization capabilities of each model on test data generated as the training ones and we measure generalization error comparing $\eta_{true}$ (\ref{eqn:innovation_moments}) with $\hat{\eta}$ where $\hat{\eta}^2:= \frac{1}{N} \sum_{i=1}^N (y_i - \hat{y}_i)^2 $ for each system.  

\begin{figure}
    \hspace{-0.5cm}
    \includegraphics[width=10cm]{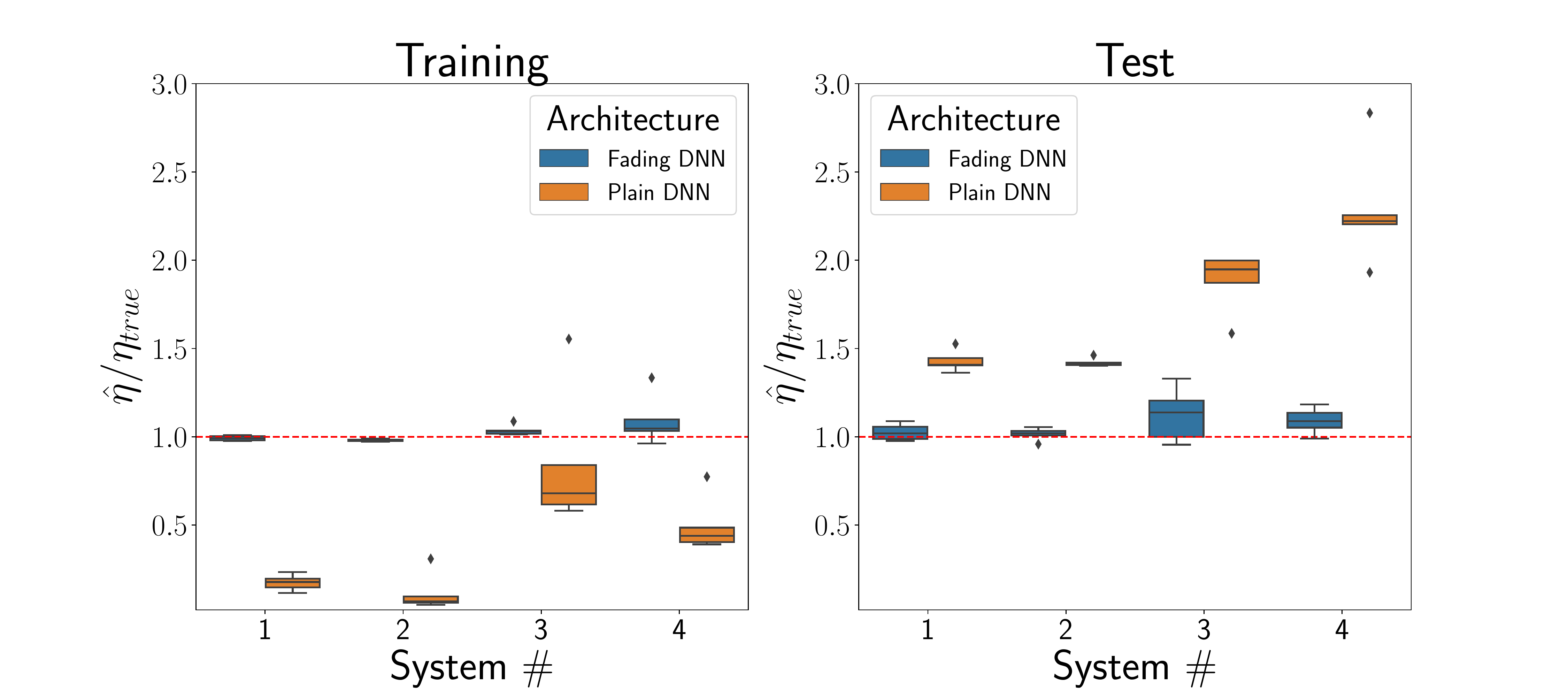}
    \caption{\textbf{Fading architecture vs plain DNN model.} Monte Carlo results: box plot for train and generalization on systems from Table \ref{tb:nlsys}  (20 runs, N=10k). Both architectures have the same input horizon (12), activations (Tanh), hidden layers (5) and a similar number of parameters. Note fading architecture avoids overfitting and reduce generalization gap for every benchmark system.}
    \label{fig:plain_vs_fading}
\end{figure}

In each experiment we choose to parametrize the blocks $f_{\mathbf{W}_i}$ using over-parametrized DNNs: 5 hidden layers, 100 hidden units with Tanh activation function ($\approx 41k$ parameters). In such a scenario we expect that without any regularization severe overfitting occurs. In Fig. \ref{fig:plain_vs_fading} we show this is indeed the case and compare a plain DNN (without any particular structure) against our fading architecture. Both models take the same number of data as input and have a similar number of parameters.

\begin{figure}[t]
    \hspace*{-0.5cm}
    \includegraphics[width=10cm]{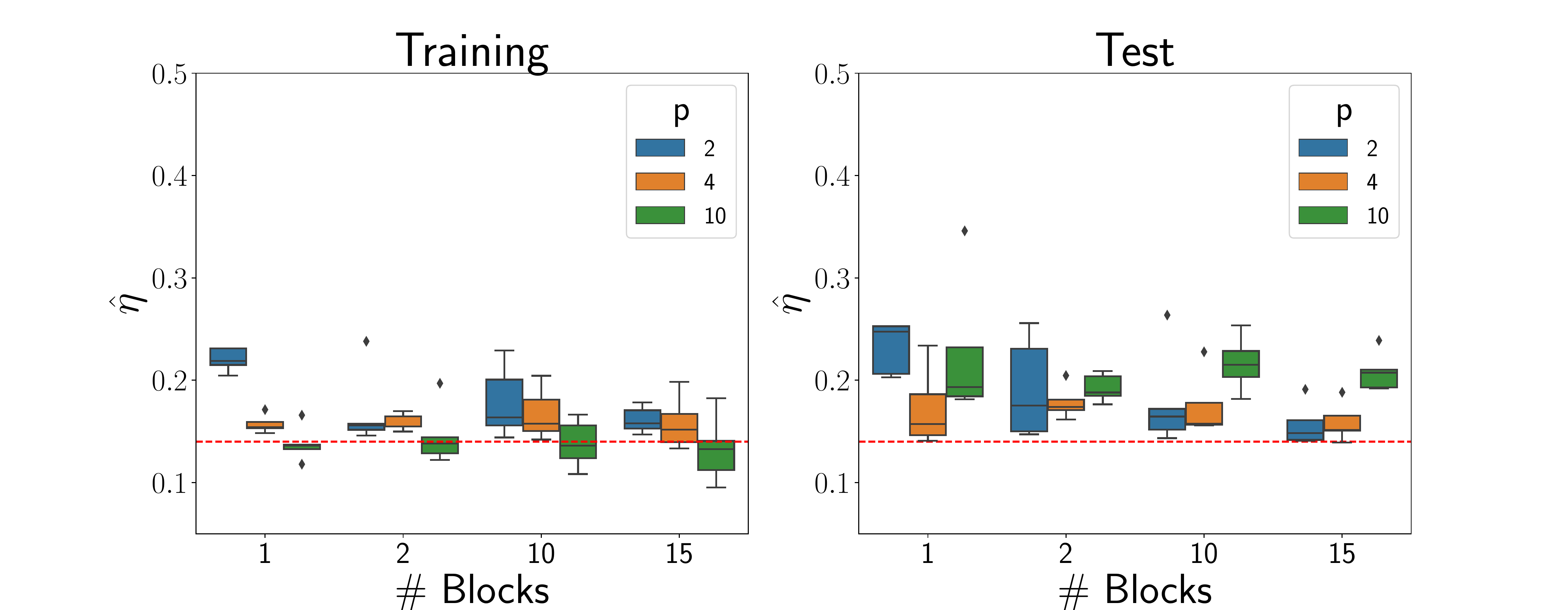}
    \hspace*{-0.5cm}
        \includegraphics[width=10cm]{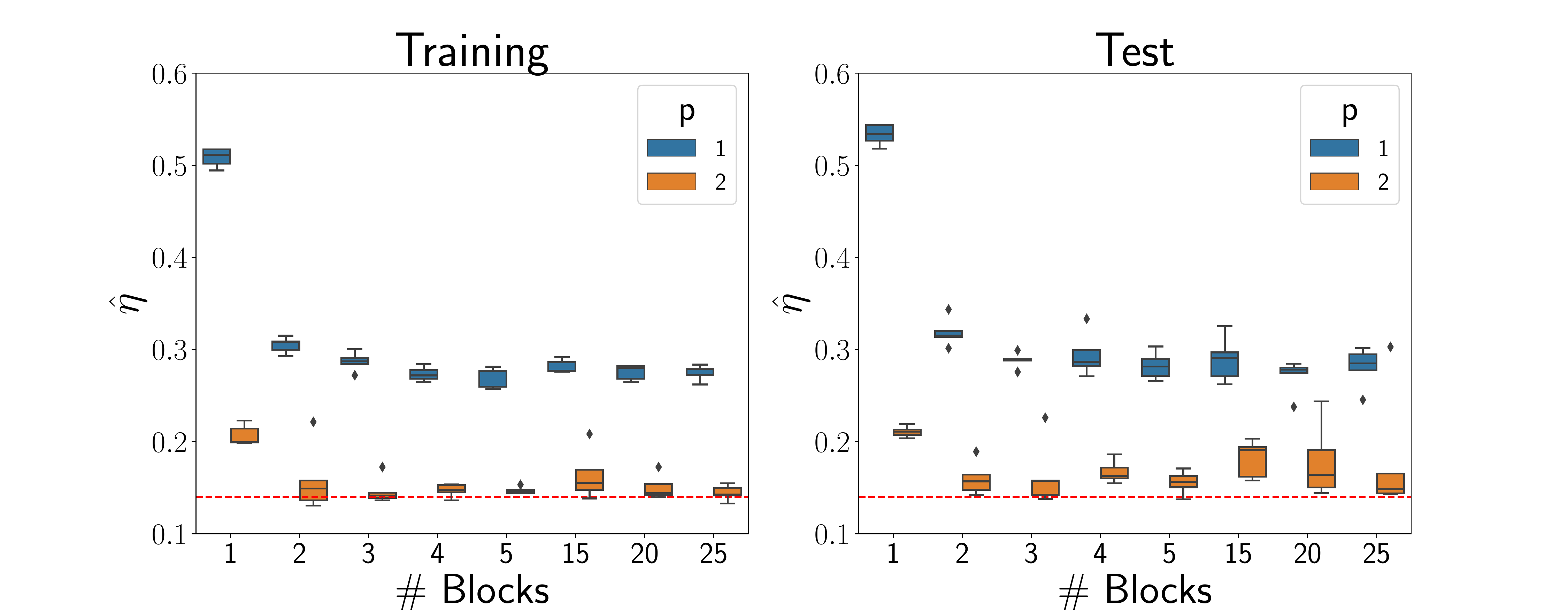}
        \caption{\textbf{Robustness of our method to the choice of horizon.} Monte Carlo results on system 4 (runs=20, N=10k) for different values of $n_B$ and $p$. 
        \textbf{Upper panels}: When $p$ is such that a single block does not overfit, our method prevents overfitting as $n_B$ grows. 
        \textbf{Lower panels}: Degenerate choice of $p$: when $p$ is too small it introduces a bias in the estimation. In this particular case we are not able to model mixed terms such as $u_{t-2}u_{t-3}$ which are present in system 4.
        }
        \label{fig:different_horizons}
\end{figure}

The proper fading horizon length is not known a priori: we tested automatic complexity selection in Fig. \ref{fig:different_horizons}. We compare different architectures  optimized according to \eqref{eqn:MAP_rewritten} using different number of blocks and block horizons $p$. We show that generalization for fixed $p$ does not worsen as the number of blocks (and therefore representational capability) increases. The robustness on the choice of the number of blocks $n_B$ proves the effectiveness of our regularization scheme. Moreover from the user's perspective it reduces the sensitivity of the identified model w.r.t. a wrong choice of the input horizon and allows the user to safely choose large $n_B$ without incurring overfitting.
Regarding the actual value of $n_B$ we have no other prescription than choosing it large enough so that the relevant past is processed by the architecture since automatic complexity selection will select $\lambda$ (the relevant past) based on available data.

One last question remains open: how to choose the horizon of each block $p$? 
Other than trial and error, cross validation could be used to choose the best hyper-parameter $p$.
In Fig. \ref{fig:different_horizons}  we compare the effects of different $p$: our regularization does not impose fading constraints on the input of each block, we therefore expect that large $p$ (despite SO regularization) might overfit. 
From the user's perspective the choice of $p$ should be as small as possible without introducing too modeling bias on each block (see Fig \ref{fig:different_horizons} for an example of a degenerate choice: $p=1$). 

\begin{figure}
    \centering
    \includegraphics[width=9cm]{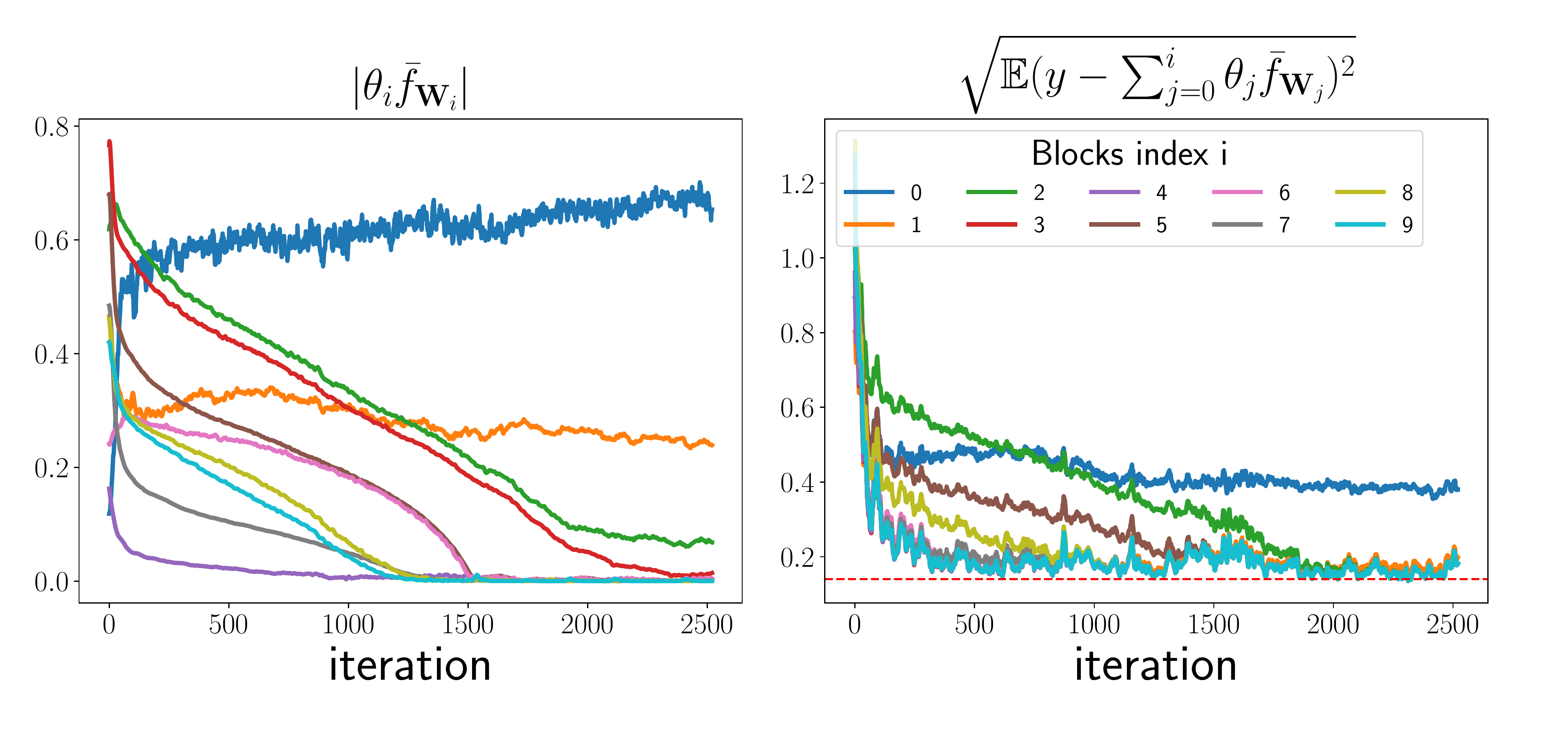}
    \caption{\textbf{Blocks' relative importance.} Single run on system 4, $n_B=9$ and $p=2$. Importance is measured both by $|\theta_i \bar{f}_{\mathbf{W}_i}|$ $i=0,...,n_B$ (\textbf{left}) and by the prediction error standard deviation of the truncated predictor up to the $i$-th block: $\sqrt{\mathbb{E}(y - \sum_{j=0}^i \theta_j \bar{f}_{\mathbf{W}_j})^2}$ for $i=0,...,n_B$ (\textbf{right}).}
    \label{fig:block_relevance}
\end{figure}

For the sake of completeness in Fig. \ref{fig:block_relevance} we show the importance of each block on the prediction $\hat{y}_t$ during optimization. We use $|\theta_i \bar{f}_{\mathbf{W}_i}|$ and the residual error of the truncated (in the number of blocks $n_B$) predictor. The latter is measured by an empirical estimate of $\sqrt{\mathbb{E}(y - \sum_{j=0}^i \theta_j \bar{f}_{\mathbf{W}_j})^2}$ for $i=0,...,n_B$. In Fig. \ref{fig:block_relevance} the block processing data closer to the present is indeed the one which mostly affects $\hat{y}_t$. Note the convergence of each block's relevance to its asymptotic value is not uniform across different blocks: the farther into the past the fastest to converge (and become negligible). We leave to future work the design of optimization schemes which could improve convergence speed (e.g. using adaptive learning rates algorithms other than Adam and other stochastic optimization methods designed to improve DNN convergence and generalization).

In Tab. \ref{tb:data_efficiency} we directly compare our architecture with the GP solution proposed in \cite{giapi}. We use system 4 to generate datasets of increasing length (up to 100k data in which case GPs cannot be used without approximation schemes). 
Our architecture shows a larger generalization gap in the low data regime but achieves increasingly better results as the dataset size increases.

\begingroup
\setlength{\tabcolsep}{3pt} %
\renewcommand{\arraystretch}{1.5} %
\begin{table}[ht]
\caption{\textbf{Data efficiency.} Comparison among: (a) GP model from \cite{giapi}, (b) Our architecture w/o SO regularization, (c) Our complete architecture. $\hat{\eta}$ median value on Monte Carlo study on system 4 ($\eta_{true}=0.14$).}\label{tb:data_efficiency}
\begin{center}
\begin{tabular}{l|cc|cc|cc|cc} \hline
    & \multicolumn{2}{c}{N=400} & \multicolumn{2}{c}{N=1000} & \multicolumn{2}{c}{N=10k} & \multicolumn{2}{c}{N=100k}\\ \hline
    & Train & Test              & Train  & Test         & Train  & Test  & Train  & Test   \\ 
(a)  & 0.14  & 0.27              &  0.13      &   0.19                &   0.14       & 0.17     &    -    &  - \\
(b)   & 0.02         & 0.49      & 0.03       &  0.45                &   0.07       &  0.23     &    0.12     &  0.20  \\
(c) & 0.10  & 0.32              &  0.15      &   0.22                &   0.16         &  0.17    &    0.15    &  0.15 \\
\hline
\end{tabular}
\end{center}
\label{tab:multicol}
\end{table}
\endgroup

\section{Conclusion}
We showed that overparametrized DNNs without a proper inductive bias and regularization fail to solve non-linear system identification benchmarks. We overcome such a limitation introducing both a new architecture inspired by fading memory systems and a new regularized loss inspired by Bayesian arguments which in turn allows for automatic complexity selection based on the observed data. 
We showed when DNN based parametric architectures are good alternatives to state of the art non-parametric models for modelling non-linear systems (mid-large data regime). 
Moreover we proved our method does not suffer from typical non-parametric models limitations on large dataset sizes and favourably scales with the number of samples.

\bibliography{ifacconf}             %

\appendix
\end{document}